\documentclass{article}

\PassOptionsToPackage{numbers, compress}{natbib}

\usepackage[dblblindworkshop,final]{neurips_2026}

\workshoptitle{Beyond Private Training:
The New Landscape of AI Privacy}

\usepackage[utf8]{inputenc} 
\usepackage[T1]{fontenc}    
\usepackage{hyperref}       
\usepackage{url}            
\usepackage{booktabs}       
\usepackage{amsfonts}       
\usepackage{amsmath}        
\usepackage{nicefrac}       
\usepackage{microtype}      
\usepackage{xcolor}         
\usepackage{graphicx}       
\usepackage{capt-of}
\usepackage{tikz}
\usetikzlibrary{positioning, arrows.meta, fit, calc, backgrounds}
\definecolor{gsInk}{HTML}{2B2B2B}
\definecolor{gsDataF}{HTML}{ECEFF3}\definecolor{gsDataL}{HTML}{5A6472}
\definecolor{gsDetF}{HTML}{DCE9F8}\definecolor{gsDetL}{HTML}{2F6FB5}
\definecolor{gsLlmF}{HTML}{FCEBD0}\definecolor{gsLlmL}{HTML}{B4740F}
\definecolor{gsOutF}{HTML}{DCEFE0}\definecolor{gsOutL}{HTML}{2E7D4F}        

\newif\ifshowcomments
\showcommentsfalse
\ifshowcomments
  \newcommand{\jiayu}[1]{\textcolor{red}{[\textit{Comment: #1}]}}
  \newcommand{\zzl}[1]{\textcolor{blue}{[\textit{Comment: #1}]}}
\else
  \newcommand{\jiayu}[1]{}
  \newcommand{\zzl}[1]{}
\fi

\title{Synthesis Without Training:\\An Inference-Only Pipeline for Tabular,\\Temporal, and Relational Synthetic Data}

\author{
  \textbf{Zilong Zhao}$^{2,3}$ \quad
  \textbf{Abdul Raheem}$^{2}$ \quad
  \textbf{Jiayu Li}$^{1}$ \\[0.4ex]
  \textbf{Sohei Arisaka}$^{4}$ \quad
  \textbf{Darius Lim Hong Yi}$^{4}$ \quad
  \textbf{Milad Abdollahzadeh}$^{2}$ \\[0.4ex]
  \textbf{Uzair Javaid}$^{2}$ \quad
  \textbf{Biplab Sikdar}$^{3}$ \\[0.6ex]
  $^{1}$University of Illinois Urbana-Champaign \\[0.2ex]
  $^{2}$Betterdata AI\quad
  $^{3}$National University of Singapore \\[0.6ex]
  $^{4}$KAJIMA Technical Research Institute Singapore
}

\begin{document}

\maketitle

\begin{abstract}
Synthetic data generation is dominated by the \emph{fit-then-sample} paradigm: a generative model is trained on a private dataset and then sampled from. Despite its widespread adoption, this paradigm faces three challenges: (1) a new training run is required for every dataset; (2) different data modalities, such as single tables, time series, and relational databases, require task-specific models and feature engineering; and (3) the resulting model is opaque, making its behavior under data constraints difficult to inspect. We propose \textsc{GenScript}, an \emph{inference-only} pipeline that eliminates model training. \textsc{GenScript} computes a deterministic statistical profile of the source data (column types, ranges, missingness, categories, correlations, etc.) and passes it---rather than raw rows---to a language model to infer field semantics and cross-column integrity constraints. A coding agent then compiles the profile and constraints into an executable, auditable sampler. This unified approach supports single-table, temporal, and relational data without task-specific modeling. Across four single-table benchmarks, \textsc{GenScript} builds generators in 2 minutes and samples 50k rows within 6 seconds, while remaining within a few points of leading methods in marginal fidelity. Notably, it is the only method that perfectly preserves a 1-to-1 mapping between columns in the Adult dataset. On a smart-building dataset, it produces conditional time series that more closely match the real distribution than two baselines and perfectly preserves primary- and foreign-key relationships in the corresponding relational database.
\end{abstract}

\section{Introduction}
\label{sec:intro}

Organisations holding sensitive tabular data increasingly rely on synthetic
surrogates for sharing, testing, and downstream model development. The standard
recipe is to fit a deep generative model to the private data and sample from it.
Conditional GANs and VAEs for tables \citep{xu2019ctgan}, diffusion models
\citep{kotelnikov2023tabddpm, shi2025tabdiff}, and large-language-model-based generators
\citep{borisov2023great} all follow this recipe, as do the specialised architectures
used for sequential \citep{yoon2019timegan, lin2020doppelganger} and relational data
\citep{patki2016sdv, solatorio2023realtabformer}.

This paradigm imposes a structural tax. \textbf{Per-dataset training} means every new
dataset needs its own run, hyperparameter search, and GPU budget, so cost scales
linearly rather than amortising. \textbf{Modality fragmentation} means a single table, a
time series, and a relational database are served by three model families with three
training procedures and three sets of feature engineering, making model \emph{selection} the
practitioner's first task. \textbf{Opacity} means that when a generator violates a
business rule---a refund on an unpaid order---there is no direct place to inspect or
repair it.

\paragraph{Our position.} For a large and practically important class of datasets,
the generative model is unnecessary. What a practitioner needs is an accurate
description of the data's marginal and joint structure, and of the rules it obeys.
Both can be obtained \emph{without} gradient descent: the first from deterministic
profiling code, the second from a language model reasoning over that profile. A coding
agent compiles both into an ordinary program that emits rows. We call this
\emph{inference-only} generation: nothing is trained at any point, and the only
learned component is a frozen, general-purpose LLM used at inference time. We instantiate this as \textsc{GenScript} (Section~\ref{sec:method}), which contributes
a \textbf{training-free pipeline} replacing the trained generator with an LLM-authored
sampling program; \textbf{one pipeline across modalities}, since the data--generator
interface is a profile rather than a tensor, so modality enters only as extra profile
fields (Section~\ref{sec:modalities}); and \textbf{explicit constraints}, materialised
as readable predicates that can be audited, edited, or supplied by an expert. We claim no advantage on data whose value lies in high-order structure a profile cannot
summarise; our claim is that for schema-driven operational data, an explicit program offers substantial reductions in computational cost and improved auditability, at a measurable cost in fidelity.

\jiayu{I think you missed some important advantages that are particularly advantages of using LLMs: 1. Not restricted by ``noisy data'' (e.g., missing data, textual fields [but not different long textual remark for each single row if script only generator], etc. that LLMs naturally understand and accommodate); 2. Making sense of the dataset's realistic meaning in real life that humans do simply but existing deep generative nets hardly do; 3. Explainability of the generation process if we're not doing rejection sampling only; 4. Easy update to any best LLM. And again if not rejection sampling only another advantage is lightning fast inference too.}

\jiayu{A better title hence may be: Making Sense Instead of Training: Synthetic Tabular Data Generation from LLM-crafted Scripts. And we may borrow the arguments for world model vs LLM.}
\vspace{-0.1em}
\paragraph{Related work.} Tabular generators fit a model to the private
data \citep{xu2019ctgan, zhao2021ctab,kotelnikov2023tabddpm, zhao2024ctab,shi2025tabdiff, li2025tabtreeformer,
borisov2023great, zhao2025tabula, solatorio2023realtabformer,zhao2023fct}, and sequential and relational data each
bring their own families again \citep{yoon2019timegan, lin2020doppelganger,
tiwald2025tabularargn, suh2025timeautodiff, pang2024clavaddpm, li2026irg}. Closest to us
are methods that use an LLM to read a table's meaning and then hand generation to a
fitted model: LLM-TabFlow \citep{long2025llmtabflow} recovers inter-column logical
relations for a diffusion model, and SPADA \citep{yang2025spada} induces a sparse
dependency graph and samples by kernel density estimation. We differ in handing
generation to a \emph{program}, so no density is estimated at any point. More  discussions are provided in Appendix~\ref{app:rw}.

\section{Method}
\label{sec:method}
\vspace{-0.2em}

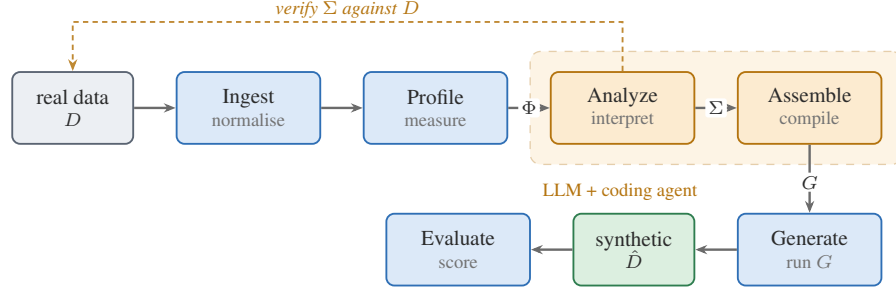
\begin{figure}[t]
  \centering
  \begin{tikzpicture}[
      font = \footnotesize,
      node distance = 9mm and 5.5mm,
      base/.style  = {rounded corners=3pt, minimum height=9.5mm, minimum width=19mm,
                      align=center, inner sep=2pt, line width=0.7pt, text=gsInk},
      det/.style   = {base, draw=gsDetL,  fill=gsDetF},
      llm/.style   = {base, draw=gsLlmL,  fill=gsLlmF},
      data/.style  = {base, draw=gsDataL, fill=gsDataF, minimum width=16mm},
      res/.style   = {base, draw=gsOutL,  fill=gsOutF, minimum width=16mm},
      sub/.style   = {font=\scriptsize, text=black!55},
      art/.style   = {font=\scriptsize\itshape, text=gsInk, fill=white,
                      inner sep=1.2pt, rounded corners=1pt},
      ar/.style    = {-{Stealth[length=4.5pt, width=3.5pt]}, line width=0.9pt, gsInk!70},
      vfy/.style   = {-{Stealth[length=4.5pt, width=3.5pt]}, line width=0.7pt,
                      dashed, dash pattern=on 2pt off 1.6pt, gsLlmL!85}
    ]
    \node[data]              (D)   {real data\\[-1pt]{\scriptsize$D$}};
    \node[det, right=of D]   (ing) {Ingest\\[-1pt]\textcolor{black!55}{\scriptsize normalise}};
    \node[det, right=of ing] (pro) {Profile\\[-1pt]\textcolor{black!55}{\scriptsize measure}};
    \node[llm, right=of pro] (ana) {Analyze\\[-1pt]\textcolor{black!55}{\scriptsize interpret}};
    \node[llm, right=of ana] (asm) {Assemble\\[-1pt]\textcolor{black!55}{\scriptsize compile}};

    \draw[ar] (D)   -- (ing);
    \draw[ar] (ing) -- (pro);
    \draw[ar] (pro) -- node[art] {$\Phi$} (ana);
    \draw[ar] (ana) -- node[art] {$\Sigma$} (asm);

    \node[det, below=of asm] (gen) {Generate\\[-1pt]\textcolor{black!55}{\scriptsize run $G$}};
    \node[res, left=of gen]  (Dh)  {synthetic\\[-1pt]{\scriptsize$\hat{D}$}};
    \node[det, left=of Dh]   (ev)  {Evaluate\\[-1pt]\textcolor{black!55}{\scriptsize score}};

    \draw[ar] (asm) -- node[art, pos=0.55] {$G$} (gen);
    \draw[ar] (gen) -- (Dh);
    \draw[ar] (Dh)  -- (ev);

    \draw[vfy] (ana.north) -- ++(0,6.5mm) -| (D.north);
    \node[font=\scriptsize\itshape, text=gsLlmL]
      at ($(D.north)!0.5!(ana.north) + (0,8.2mm)$) {verify $\Sigma$ against $D$};

    \begin{scope}[on background layer]
      \node[rounded corners=4pt, fill=gsLlmF!55, draw=gsLlmL!35, dashed,
            line width=0.5pt, inner sep=2.6mm, fit=(ana)(asm)] (band) {};
    \end{scope}
    \node[sub, text=gsLlmL, anchor=north west] at ([xshift=0.5mm, yshift=-1.2mm]band.south west) {LLM + coding agent};
  \end{tikzpicture}
  \caption{The \textsc{GenScript} working pipeline. The implemented UI is provided in Appendix~\ref{app:ui}.
  }
  \label{fig:pipeline}
  \vspace{-1.5em}
\end{figure}

Let $D = \{T_1,\dots,T_K\}$ be a source dataset of $K$ tables ($K=1$ for the
single-table case). \textsc{GenScript} produces $\hat{D}$ through the six phases of
Figure~\ref{fig:pipeline}: \emph{Ingest}, \emph{Profile}, \emph{Analyze} and \emph{Assemble}
construct the generator, \emph{Generate} runs it, and \emph{Evaluate} scores the output. 

\subsection{Ingest and Profile: deterministic statistical profiling}
\vspace{-0.1em}
\label{sec:profiling}

A fixed, hand-written script computes a profile $\Phi(D)$; no language model is
involved, so this phase is exact, cheap, and reproducible. Per column we record the
inferred type (continuous, integer, categorical, datetime, text, identifier); the
observed range, or for categorical columns the category inventory with empirical
frequencies; the missing-value rate and any conditional missingness; and summary
statistics. Per table we compute a correlation matrix; for
relational inputs we record keys and each parent--child cardinality distribution; for time series the profile adds the timestamp granularity and the distribution of inter-arrival gaps, per-entity sequence-length and coverage statistics.

\jiayu{I think it is also very helpful if we obtain some regression analysis keeping any single column as the output, sometimes even masking out very highly correlated columns when that correlation is already included in the information}
\jiayu{I doubt this, as I think stage 2 may still allow LLM agents to run scripts on raw data, but maybe just not showing concrete values from the raw data.}

\jiayu{One thing in addition: only statistical profiling? I think textual description of what the columns are and dataset is is also essential. Make use of the advantage of LLMs!}

\subsection{Analyze: semantic analysis and constraint induction}
\vspace{-0.1em}
\label{sec:semantics}

We pass $\Phi(D)$, including the correlation matrix, to a language model and ask
for a semantic specification $\Sigma$. This recovers information present in the data but
invisible to the profiler: \textbf{field semantics}, such as that a five-digit integer
column is a postal code rather than a quantity; \textbf{cross-column constraints} such as
``if \texttt{A}~$>0$ then \texttt{B}~$=0$'', for which strong correlation entries and
structured missingness are evidence that a rule exists while column names supply its
form; and \textbf{conditional structure}, indicating which columns to sample conditioned
on which others.
\jiayu{There is missing column wise scripts, e.g., two binary columns actually better understands as 1-column 4 categories, then its distributino? and whether the column follow X distribution (instead of enumerating during Stage 1, especially for non normal distributions like lognormal, exponential)}

Formally $\Sigma = \{(c_j, w_j)\}$ is a set of predicates over a row---or a row and its
parent, in the relational case---with confidence weights, short enough to be reviewed and
corrected by a domain expert before generation.
Every candidate predicate is additionally \emph{verified against the real data}:
those holding on fewer than $\tau$ of real rows are demoted or discarded, guarding
against rules hallucinated from suggestive column names alone.
\jiayu{Similarly to above, maybe not just row level. And I think this is where we can firmly say that no touch on real data from this point on. Stage 2 is only the end of DIRECT visibility but indirect is still there}

\subsection{Assemble, Generate and Evaluate: synthesizing and running the sampler}
\label{sec:agent}
\vspace{-0.1em}

A coding agent receives $\Phi$ and $\Sigma$ and writes an executable generator $G$,
given a target interface and a sandbox in which to run its own output. Generation
proceeds by constrained rejection sampling: the program draws a candidate row from a
factorized proposal built from the profile's marginals and conditional structure,
accepting it only if the hard predicates hold,\jiayu{Again I am a bit doubtful of whether we want rejection sampling only, we initially turn to this proposal wanting to get rid of the cost of sampling but now you seem not really tackling the issue; script only generation would be lightning fast but current sampling speed isn't fast at all. Moreover, I do not agree that reject sampling is distribution-preserving because some conditions naturally centralize on certain column values and usually gets rejected because the constraint is hard, then the overall distribution is distorted. It is ok to use some rejection pipeline, but rejection to force the agent to edit the program is better.}
\begin{equation}
  \hat{x} \sim q_{\Phi}(\cdot), \qquad
  \text{accept if } \textstyle\prod_{j \in \mathcal{H}} c_j(\hat{x}) = 1 ,
\end{equation}
where $\mathcal{H}$ indexes the hard constraints\jiayu{Again derived from my previous comment, constraint may include multi row ones on the distribution}. Where a constraint is
deterministic---an arithmetic identity, a derived field---the agent \emph{imputes}
rather than rejects\jiayu{The word ``impute'' has specific meaning, I think it's not a correct word here}, computing the dependent value directly; rejection is reserved for
inequality-shaped constraints, which keeps acceptance rates high. 
Eventually, synthetic data $\hat{D}$ are evaluated as shown in Section~\ref{sec:experiments}.

\subsection{Extension to time series and relational data}
\label{sec:modalities}
\vspace{-0.1em}
\jiayu{Just a general comment, I feel the contribution is substantial enough that we may not want to squeeze such topic into the same paper here. They do require additional scripting to handle additional relations that can lead to new papers.}

The modalities differ only in what \emph{Profile} records and what \emph{Generate}
emits. For time series the profile adds the sampling interval, per-entity
sequence-length distribution, trend and seasonality estimates, and lag-$k$
autocorrelations. For relational databases it adds the schema graph and the distribution
of child-row counts per parent, and the agent walks the schema in topological order with
foreign keys assigned by construction, so referential integrity holds by design rather
than by filtering. The practitioner selects no model, because there is none to select.

\section{Experiments}
\label{sec:experiments}
\vspace{-0.2em}

\paragraph{Setup.} For single tables we use CoverType, Credit, Intrusion, and Adult,
covering heavy categorical structure, class imbalance, and large row counts; per-dataset
statistics for these and for the two relational tables are given in
Appendix~\ref{app:datasets}. Baselines are
TabTreeFormer \citep{li2025tabtreeformer}, TabDiff \citep{shi2025tabdiff}, and
REaLTabFormer \citep{solatorio2023realtabformer}, one from each of the tree-based,
diffusion, and auto-regressive language-model families. Each method generates as many rows
as the source table contains: 50\,000 for the first three, 32\,561 for Adult. $\tau$ is set to 100 rows for the semantic analysis phase. \textsc{GenScript} uses gpt-5.6-sol as its LLM
backend; the baselines were trained on a server with 32 CPU cores, 200\,GB RAM, and two
RTX 4090 Ti GPUs. 
\jiayu{Also I think it's necessary to include generators that are inference only because they do exist. Also necessary to include statistical only models cuz we use LLMs to generate statistical scripts.}

\vspace{-0.1em}
\paragraph{Metrics.} Fidelity uses \emph{Shape} and \emph{Trend}
\citep{sdmetrics2023, shi2025tabdiff}: Shape is the similarity of each column's marginal
density, Trend the fidelity of correlations between column pairs; higher is better for
both. Utility uses machine-learning efficacy under train-on-synthetic, test-on-real: we
split the real data $8\!:\!2$, fit an XGBoost classifier on the synthetic table, and
report AUC on the held-out real $20\%$. We also report generator-construction
(``Train'') and sampling time; for \textsc{GenScript}, ``Train'' covers
profiling, constraint induction, and program synthesis, with no gradient step in it.

\begin{table}[t]
  \caption{Fidelity and utility on the four single-table benchmarks. 
  Best per column in bold.}
  \label{tab:single}
  \centering
  \footnotesize
  \setlength{\tabcolsep}{3pt}
  \begin{tabular}{lcccccccccccc}
    \toprule
    & \multicolumn{3}{c}{CoverType} & \multicolumn{3}{c}{Credit} & \multicolumn{3}{c}{Intrusion} & \multicolumn{3}{c}{Adult} \\
    \cmidrule(lr){2-4} \cmidrule(lr){5-7} \cmidrule(lr){8-10} \cmidrule(lr){11-13}
    Method & Shape & Trend & MLE & Shape & Trend & MLE & Shape & Trend & MLE & Shape & Trend & MLE \\
    \midrule
    TabTreeFormer      & \textbf{0.991} & \textbf{0.991} & 0.919          & 0.929          & 0.942          & 0.971          & 0.866          & 0.982          & 0.565          & 0.935          & 0.972          & 0.822 \\
    REaLTabFormer      & 0.983          & 0.990          & \textbf{0.937} & 0.962          & 0.948          & 0.947          & 0.968          & 0.962          & \textbf{0.999} & 0.966          & 0.930          & \textbf{0.925} \\
    TabDiff            & 0.975          & 0.820          & 0.894          & \textbf{0.992} & \textbf{0.995} & \textbf{0.999} & \textbf{0.989} & \textbf{0.989} & 0.995          & \textbf{0.986} & \textbf{0.973} & 0.924 \\
    \textsc{GenScript} & 0.984          & 0.784          & 0.467          & 0.930          & 0.897          & 0.894          & 0.874          & 0.748          & 0.572          & 0.936          & 0.858          & 0.828 \\
    \bottomrule
  \end{tabular}
\end{table}

\begin{table}[t]
  \caption{Generator-construction time (``Train'') and sampling time, in
  seconds. For \textsc{GenScript}, ``Train'' covers profiling, constraint induction, and
  program synthesis.}
  \label{tab:time}
  \centering
  \footnotesize
  \setlength{\tabcolsep}{4pt}
  \begin{tabular}{lcccccccc}
    \toprule
    & \multicolumn{2}{c}{CoverType} & \multicolumn{2}{c}{Credit} & \multicolumn{2}{c}{Intrusion} & \multicolumn{2}{c}{Adult} \\
    \cmidrule(lr){2-3} \cmidrule(lr){4-5} \cmidrule(lr){6-7} \cmidrule(lr){8-9}
    Method & Train $\downarrow$ & Sample $\downarrow$ & Train $\downarrow$ & Sample $\downarrow$ & Train $\downarrow$ & Sample $\downarrow$ & Train $\downarrow$ & Sample $\downarrow$ \\
    \midrule
    TabTreeFormer      & 2324.9         & 474.9        & 273.6         & 63.1         & 217.5          & 51.8         & 2117.0        & 43.6 \\
    REaLTabFormer      & 405.9          & 159.6        & 2194.5        & 533.5        & 581.0          & 249.9        & 175.1         & 37.8 \\
    TabDiff            & 5893.7         & 16.5         & 4934.3        & 13.1         & 5567.1         & 16.1         & 6760.0        & 8.6 \\
    \textsc{GenScript} & \textbf{112.2} & \textbf{3.0} & \textbf{93.0} & \textbf{5.4} & \textbf{114.9} & \textbf{2.5} & \textbf{73.3}          & \textbf{0.9} \\
    \bottomrule
  \end{tabular}
\end{table}

\paragraph{Fidelity and utility.} On Shape \textsc{GenScript} is close to the leaders
(0.984, 0.936, 0.930 and 0.874 on CoverType, Adult, Credit and Intrusion), second of four
on CoverType and third elsewhere, which is a good outcome for a sampler built from
per-column marginals alone. On Trend it is last on all four (0.748 to 0.897), the direct
cost of reproducing only the dependencies named in $\Sigma$. MLE is the weakest axis and
degrades with the number of target classes: 0.894 on Credit and 0.828 on Adult, both
binary, against 0.572 on Intrusion and 0.467 on CoverType, which have twenty and seven
classes respectively. Further discussions are provided in Appendix~\ref{app:results}.


\vspace{-0.1em}
\paragraph{Constraint preservation.} The aggregate metrics are blind to hard integrity
rules. In Adult, \texttt{education} and \texttt{education-num} encode the same fact twice,
so the real table admits exactly 16 diploma/year pairs. Every trained baseline emits rows
violating this dependency, whereas \textsc{GenScript} materialises it as a hard predicate
and emits none. Such rows are impossible rather than improbable, and no metric in
Table~\ref{tab:single} registers them; Appendix~\ref{app:fd} gives per-method rates.

\vspace{-0.1em}
\paragraph{Cost.} Table~\ref{tab:time} is where the inference-only design pays.
\textsc{GenScript} is the fastest method among all the baselines. It builds its
generator in 73 to 115 seconds, against 175 to 6760 seconds for the baselines, and samples
in 0.9 to 5.4 seconds, against 8.6 to 534 seconds. 
Measured against whichever baseline is
fastest on each dataset, construction is 1.9 to 3.6 times quicker and sampling 2.4 to 9.6
times quicker; 
End-to-end the totals are 74 to 117 seconds against 213 to 566
seconds for the best baseline on each dataset, a margin of 2.3 to 4.9 times achieved.
\jiayu{Instead of reporting train and sample times separately in the table that also confuses readers when you do seem to have some training time, I think maybe you instead model the total time training and sampling as y=kx+b where x means the number of rows sampled, and we want to show smaller k and b. }

\vspace{-0.1em}
\paragraph{Time series and relational structure.} For the remaining modalities we use a
mock-up dataset from an industrial partner in the construction sector. An access-control
table (\texttt{acc}) records everyone granted entry to a building, keyed by
\texttt{UserID}; a camera-detection table (\texttt{aicamera}) records their sightings as a
timestamped time series whose \texttt{UserID} references \texttt{acc}, so the pair is both
a conditional time series and a two-table relational database. On the relational axis
nothing separates the methods: all preserve primary-key uniqueness and referential
integrity. What differs is that the baselines must be told the key, whereas
\textsc{GenScript} identifies it from the profile alone. On the temporal axis they do
separate. The t-SNE result ( Figure~\ref{fig:tsne} in Appendix~\ref{app:tsne}) reverses the
single-table picture: \textsc{GenScript}'s points interleave with the real ones throughout
the main manifold, whereas both trained baselines form clusters entirely disjoint from it.
Two properties of \texttt{aicamera} make it well suited to an induced rule set. The
building is an office, so presence follows a strong daily routine.
And \texttt{source.id} (i.e., the time series column in \texttt{aicamera}) is categorical, so its per-category frequencies are recorded exactly in the profile and reproduced by construction. 

\section{Conclusion and future work}
\label{sec:limitations}
\vspace{-0.1em}
\textsc{GenScript} shows that a useful synthetic-data generator can be obtained without
training one. Across four single-table benchmarks it is the fastest method on both
construction and sampling while staying close to the leaders on marginal fidelity; on a
real building-telemetry time series it matches the real distribution more closely than two
specialised temporal models; and it alone emits no rows violating a functional dependency,
because the rule is named and enforced rather than fitted. What it produces is a program a
practitioner can read, edit and re-run. The limits are equally clear: it is last on Trend
on all four datasets, so label-conditional structure is not yet reproduced.

Three directions follow. The constraints we induce are essentially pairwise, since the
correlation matrix is the evidence driving them; dependencies over three or more columns
are invisible to it and are exactly what wide tables contain, so extending induction beyond
pairwise structure is the most direct route to closing the Trend gap. The profile is also
the only input to the semantic phase: a short description of the dataset would let the
model propose candidate relationships from domain knowledge, which the verification step of
Section~\ref{sec:semantics} can confirm or reject against the real data. Finally, our
relational evaluation covers one two-table schema; deeper hierarchies, composite keys and
cycles remain untested.

\small
\bibliographystyle{plainnat}
\bibliography{references}
\normalsize





\newpage
\appendix

\section{Extended related work}
\label{app:rw}

\paragraph{Related work: single tables.} CTGAN and TVAE \citep{xu2019ctgan}, the
diffusion models TabDDPM \citep{kotelnikov2023tabddpm} and TabDiff
\citep{shi2025tabdiff}, the hybrid tree-transformer TabTreeFormer
\citep{li2025tabtreeformer}, and the language-model generators GReaT
\citep{borisov2023great} and REaLTabFormer \citep{solatorio2023realtabformer} all train
on the private data. TabDiff and TabTreeFormer are the closest analogues to our aim, but
both unify \emph{within} the single-table modality and stay inside weight space, whereas
we unify \emph{across} modalities by leaving weight space altogether. PrivBayes
\citep{zhang2017privbayes} and SDV \citep{patki2016sdv} share our commitment to an
explicit model, but fit their dependency structure and cannot express logical
constraints.

\paragraph{Related work: sequential and relational.} For sequential data, TimeGAN
\citep{yoon2019timegan} and DoppelGANger \citep{lin2020doppelganger} are adversarial;
TabularARGN \citep{tiwald2025tabularargn} trains auto-regressive conditionals across the
column, time, and table axes; and TimeAutoDiff \citep{suh2025timeautodiff} pairs a VAE
with latent diffusion for heterogeneous time-series tables. For relational schemas, ClavaDDPM
\citep{pang2024clavaddpm} propagates cluster latent variables across foreign keys, and
IRG \citep{li2026irg} generates tables incrementally along a depth-first traversal to
handle composite and overlapping keys. Each is strong on its home ground, and
collectively they make our point: three modalities, three literatures, three training
procedures. Our aim is not to beat any one in its own setting, but to cover all three
with one pipeline and no training at all. Our third stage separately builds on the observation
that language models are more reliable emitting programs than answers
\citep{gao2023pal, yao2023react}: ours is never asked to produce rows, only the sampler
that produces them, which bounds its contribution to a short verifiable artifact and
makes generation deterministic given a seed.

\jiayu{Related work needed: LLMs making sense of tabular data.}

\paragraph{Related work: LLMs making sense of tabular data.} A separate line of work
uses language models not as density estimators but as readers of a table's meaning.
TabLLM \citep{hegselmann2023tabllm} shows that serialising rows into text lets a
pretrained model classify from a handful of examples, evidence that column names and
value formats carry usable semantics on their own. The data-management literature has
pushed this further: \citet{narayan2022wrangle} cast entity matching, error detection,
and imputation as prompting tasks and find that foundation models reach
state-of-the-art without task-specific training, and ArcheType
\citep{feuer2024archetype} performs zero-shot semantic column-type annotation, assigning
meaning to a column from its name and a sample of its values rather than from a fixed
type vocabulary learned in advance. Our \emph{Analyze} phase is the same operation put to a different
end: where that work annotates a column to clean, match, or integrate it, we annotate it
in order to \emph{generate} it, and we ask additionally for the constraints that hold
\emph{between} columns. LLM-TabFlow
\citep{long2025llmtabflow} uses LLM reasoning to recover inter-column logical
relationships and then delegates density modelling to a score-based diffusion model,
and SPADA \citep{yang2025spada} induces a sparse dependency graph with an LLM and
synthesises by traversing it with kernel density estimation or a normalising flow,
avoiding LLM calls at sampling time entirely. These are the closest precedents for what
we do, and they bracket our position: both extract structure with an LLM and then hand
generation to a fitted statistical model, whereas we hand it to a program the LLM
writes, so no density is estimated at any point and the extracted structure stays
legible in the artifact that generates the data. Purely statistical generators such as
PrivBayes \citep{zhang2017privbayes} and SDV \citep{patki2016sdv} sit at the other end:
inspectable and cheap, but with the dependency structure fitted rather than named, and
no mechanism for the semantics that column names carry.

\section{Dataset statistics}
\label{app:datasets}

\begin{table}[h]
  \caption{The six source tables. Columns counts include the target where one exists.
  Following common practice for these benchmarks, CoverType, Credit and Intrusion are
  subsampled to 50\,000 rows stratified on the target; Adult is used at its standard
  training-split size. Rows generated equals rows in the source table for every method.}
  \label{tab:datasets}
  \centering
  \footnotesize
  \setlength{\tabcolsep}{4pt}
  \begin{tabular}{llrrcl}
    \toprule
    Table & Domain & Rows & Cols. & Classes & Source \\
    \midrule
    CoverType & forest cover      & 50\,000 & 55 & 7 & UCI Covertype (581\,012 rows) \\
    Credit    & card transactions & 50\,000 & 31 & 2 & Kaggle creditcardfraud (284\,807) \\
    Intrusion & network traffic   & 50\,000 & 42 & 20 & UCI KDD Cup 1999 \\
    Adult     & census income     & 32\,561 & 15 & 2 & UCI Adult (48\,842 in full) \\
    \midrule
    \texttt{acc}      & building access & 100 & 6 & --- & industrial partner \\
    \texttt{aicamera} & camera detections & 138\,341 & 4 & --- & industrial partner \\
    \bottomrule
  \end{tabular}
\end{table}

The four public tables are the benchmark suite used by the CTAB-GAN line of work and its
successors, which is why the subsampling protocol follows theirs: 50\,000 rows drawn
stratified on the target for the three large tables, Adult taken as-is. Intrusion's target is the raw
KDD Cup 1999 label, not the coarse five-way grouping into normal traffic plus four attack
families that much of the intrusion-detection literature uses. The 10\%\ KDD file carries
23 distinct labels, and stratified subsampling to 50\,000 rows drops the rarest of them
(\texttt{spy}, \texttt{perl} and \texttt{phf} each occur fewer than five times), leaving
the 20 classes we synthesise. This matters for the MLE numbers in
Table~\ref{tab:single}: Intrusion is a 20-way problem, so its AUC is macro-averaged over
far more classes than CoverType's seven.
The two partner tables have no target column, so no class count is
reported for them. They are also very differently shaped: \texttt{acc} is a small
reference table of 100 rows, while \texttt{aicamera} holds 138\,341 detections, an average
of roughly 1\,400 per registered user. Its \texttt{source.id} column takes 242 distinct
values, which is the high-cardinality categorical whose frequencies the profile records
exactly and the generator reproduces by construction (Section~\ref{sec:experiments}).

\section{Per-dataset fidelity and utility}
\label{app:results}

\paragraph{Fidelity.} On Shape, \textsc{GenScript} is competitive without being best:
0.984 on CoverType places it second of four, ahead of both TabDiff and REaLTabFormer and
within 0.7 points of TabTreeFormer, and 0.936 on Adult and 0.930 on Credit sit within 5
and 6 points of the leader. This is a good outcome for a sampler built directly from
per-column marginals. Intrusion is the hardest case at 0.874, though TabTreeFormer fares
far worse there (0.557), suggesting its high-cardinality categorical columns are
difficult for any method that discretises them. On Trend, \textsc{GenScript} is last on
all four datasets (0.748--0.897). This is the cost of the design: the sampler reproduces
only the dependencies named in $\Sigma$ or carried by the proposal's conditional
structure, so pairwise correlations that no induced rule captured are not modelled. Trend
is where a trained generator has the clearest advantage, and our results do not dispute
it.
\jiayu{We mentioned the constraints, we can also create constraints and evalaute against constraints too.}

\paragraph{Utility.} MLE is where \textsc{GenScript} trails furthest, and the pattern
follows the target's cardinality. On the binary-target datasets it is usable if behind:
0.894 AUC on Credit and 0.828 on Adult, against 0.999 and 0.925 for the best baseline. On
the multi-class datasets it degrades sharply---0.572 on Intrusion and 0.467 on CoverType,
the latter still below chance, so a classifier trained on that output is worse than
useless on real data. No baseline drops below chance on any dataset, so this points at
our handling of multi-class labels rather than at a limit of the approach: the sampler
draws the target from its marginal without conditioning on the features that determine
it, which costs little when the target is binary and much more when it is not. 


\section{Implementation}
\label{app:ui}

\begin{figure}[h]
  \centering
  \includegraphics[width=0.95\linewidth]{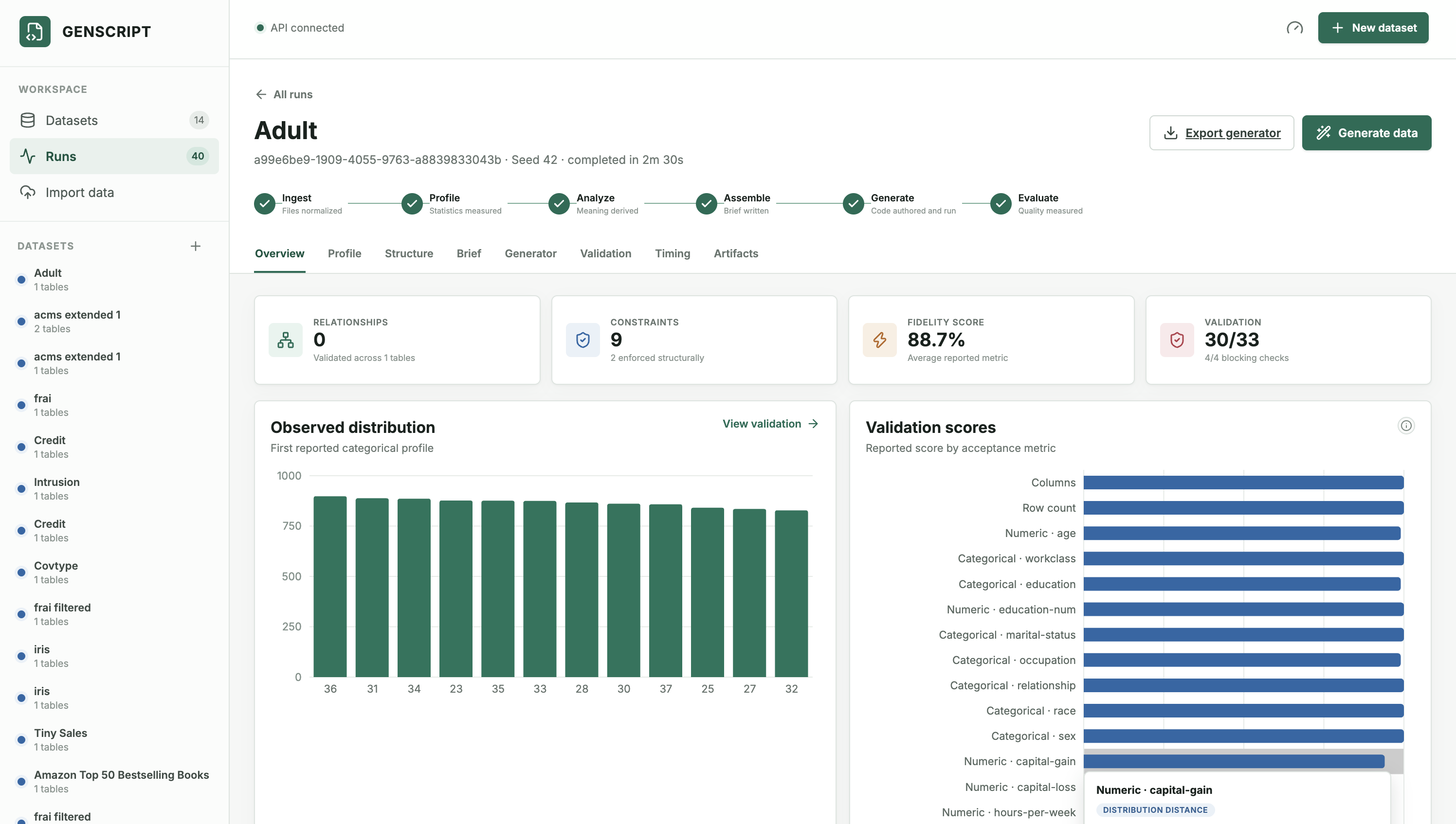}
  \caption{The \textsc{GenScript} application user interface, on a completed Adult run. The phase bar
  tracks the pipeline of Section~\ref{sec:method}. The cards report what this run
  recovered: nine induced constraints, two of them enforced structurally, and 30 of 33
  validation checks passed. The generator is exportable, and so auditable and re-runnable
  independently of the pipeline that wrote it.}
  \label{fig:ui}
\end{figure}

\section{Time-series visualisation}
\label{app:tsne}

\begin{figure}[h]
  \centering
  \includegraphics[width=0.46\linewidth]{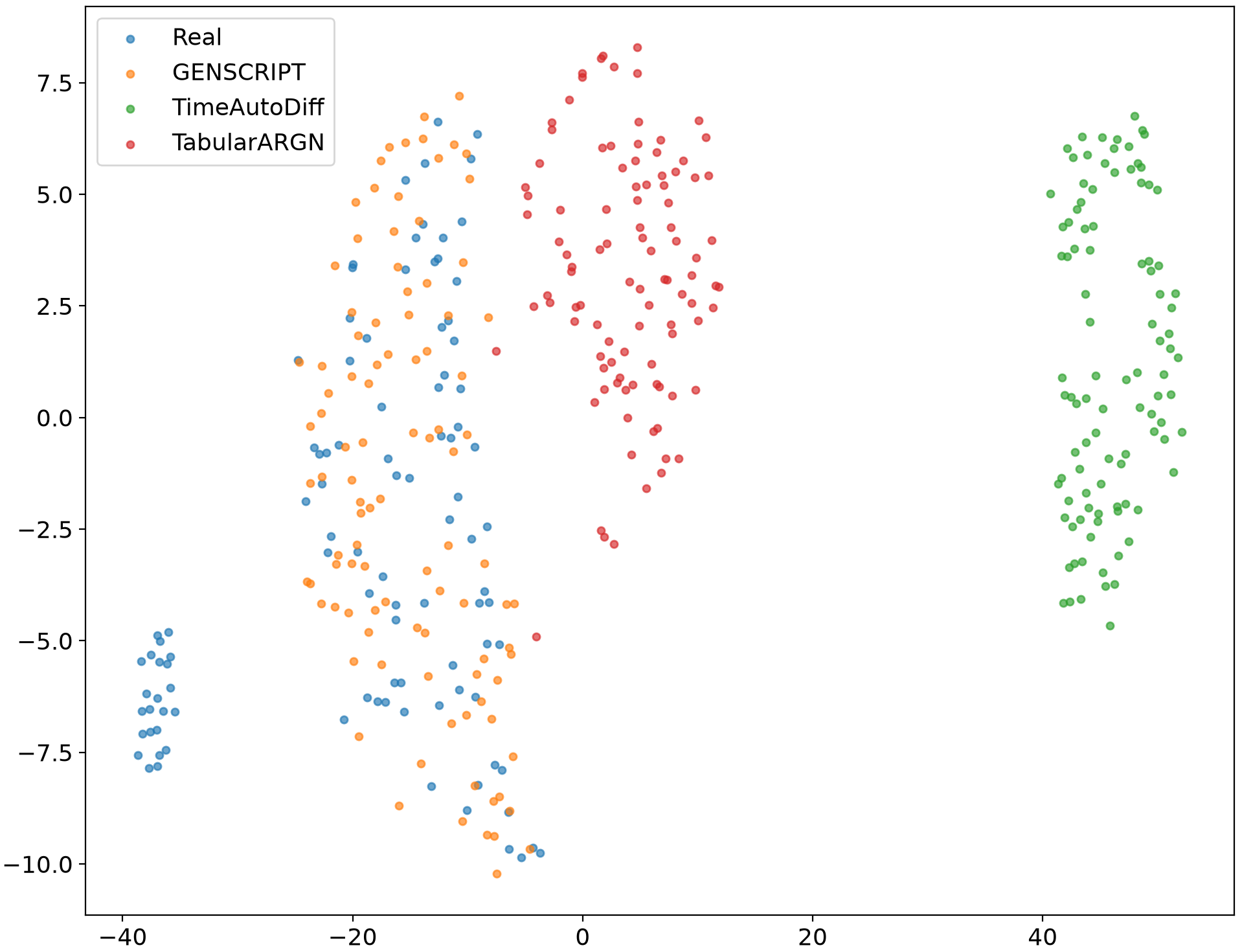}
  \caption{t-SNE of \texttt{aicamera} sequences embedded by \texttt{source.id}. \textsc{GenScript}
  (orange) lies inside the real manifold (blue); TabularARGN (red) and TimeAutoDiff
  (green) occupy disjoint regions.
  }
  \label{fig:tsne}
\end{figure}

\section{Constraint preservation on Adult}
\label{app:fd}

\begin{table}[h]
  \caption{Consistency of the \texttt{education}\,$\leftrightarrow$\,\texttt{education-num}
  functional dependency on Adult. The real table maps each of its 16 diploma labels to
  exactly one year count; a row whose year count does not match its label is a violation.
  Percentages are of the 32\,561 generated rows.}
  \label{tab:fd}
  \centering
  \footnotesize
  \begin{tabular}{lccc}
    \toprule
    Method & Distinct pairs (real: 16) & Violating rows (\%) $\downarrow$ & FD holds \\
    \midrule
    TabTreeFormer      & 16          & 61.104         & No \\
    REaLTabFormer      & 19          & 0.009          & No \\
    TabDiff            & 30          & 0.190          & No \\
    \textsc{GenScript} & \textbf{16} & \textbf{0.000} & \textbf{Yes} \\
    \bottomrule
  \end{tabular}
\end{table}


In Adult, \texttt{education} and \texttt{education-num} encode the same fact twice---a
diploma label and the year count it corresponds to---so the real table contains exactly
16 distinct pairs and the mapping is a functional dependency. Table~\ref{tab:fd} shows
that no trained baseline reproduces it exactly, and that the distinct-pair count alone is
not a sufficient diagnostic: a method can emit the right \emph{number} of pairs and still
pair a diploma with the wrong year count, so the row-level violation rate is the measure
that matters.

The violation rates are small for some baselines and large for others, but the
distinction that matters is between zero and non-zero rather than between rates. A
functional dependency admits no exceptions: a row pairing a diploma with an impossible
year count is invalid however rare it is, and rare violations are in some ways worse than
frequent ones, because they survive spot-checks and surface later in whatever downstream
job joins or filters on the column. None of these violations is visible to Shape, Trend,
or MLE.

\textsc{GenScript} recovers the dependency in \emph{Analyze} (Section~\ref{sec:semantics})---the
correlation is the evidence that a rule exists, the column names supply its form---and
materialises it as a hard predicate, so the violation count is zero by construction
rather than by filtering afterwards. This is the concrete payoff of naming constraints
instead of fitting them.

\end{document}